# PRFashion24:

# A Dataset for Sentiment Analysis of Fashion Products Reviews in Persian

Mehrimah Amirpour , reza azmi[1]

Department of Computer Engineering,Faculty of Engineering, Alzahra University, Tehran, Iran

M.amirpour@student.alzahra.ac.ir, azmi@alzahra.ac.ir

**Abstract.** The PRFashion24 dataset is a comprehensive Persian dataset collected from various online fashion stores, spanning from April 2020 to March 2024. With 767,272 reviews, it is the first dataset in its kind that encompasses diverse categories within the fashion industry in the Persian language. The goal of this study is to harness deep learning techniques, specifically Long Short-Term Memory (LSTM) networks and a combination of Bidirectional LSTM and Convolutional Neural Network (BiLSTM-CNN), to analyze and reveal sentiments towards online fashion shopping. The LSTM model yielded an accuracy of 81.23%, while the BiLSTM-CNN model reached 82.89%. This research aims not only to introduce a diverse dataset in the field of fashion but also to enhance the public's understanding of opinions on online fashion shopping, which predominantly reflect a positive sentiment. Upon publication, both the optimized models and the PRFashion24 dataset will be available on GitHub.

## 1 Introduction

Sentiment analysis is essential for organizations, particularly those in the competitive retail industry, as it allows them to precisely assess the attitudes of their customers. Due to its ability to combine natural language processing with the ability to interpret the emotional undertones of consumer comments, this technology is becoming more and more significant for fashion products on the leading online retailer in Iran [1][2]. Combining sentiment analysis in the Persian (Farsi) environment, especially for fashion items, highlights the deeper emotional ties these products engender, expressing customers' identity and mood in addition to illuminating customers' feelings, whether they be delight, irritation, or apathy [2].

This study focuses on the PRFashion24 dataset from several online stores and aims to demonstrate how complex natural language processing can understand consumer sentiments about fashion products by analyzing sentiments in Persian, a language with barriers. It is a special language, improve it. [1][2]. As a result, the article will cover a wide range of topics, including deep learning's application in sentiment analysis and data preprocessing techniques. However, it will concentrate on the nuances of processing content in Persian and the overall effects of sentiment analysis on business metrics like customer satisfaction, sales, and innovation [2].

Sentiment analysis is an effective method for figuring out how consumers feel about fashion items. Businesses can learn more about their consumers' emotional reactions by examining social media posts, reviews, and other forms of feedback. Customized marketing plans, better product designs, and higher levels of client happiness can all be achieved with the use of this data. For example, if sentiment research indicates that consumers have favorable opinions about a specific design or feature, a company can concentrate on highlighting such elements. On the other hand, if unfavorable opinions are found, a brand may be prompted to address issues or make the required changes. In the end, sentiment analysis fosters a more engaging and customized purchasing experience by assisting fashion firms in remaining aware of the demands and preferences of their customers [3].

This essay focuses on the Persian (Farsi) language, which is spoken and utilized by about 110 million people in nations including Tajikistan, Afghanistan, and Iran. Its adjacent languages, including the Turkic, Armenian, Georgian, and Indo-Aryan languages, have been greatly influenced by it. Thirty-two characters, written from right to left, make up its alphabet.

---

[1] corresponding author

Sentiment analysis is a commonly used text classification task in natural language processing that entails obtaining the sentiment that people have expressed toward a range of entities, such as goods, services, organizations, people, problems, occasions, and subjects, along with their corresponding characteristics[4]. In this case, sentiment refers to the people's positive, negative, or neutral opinions as stated in the text that was retrieved from the source, according to Jurafsky and Martin [5] . Numerous industries, including politics, healthcare, social media, marketing, and finance [6], have found extensive uses for sentiment research.

The presented dataset is extracted from several sites related to fashion and clothing. We developed a deep learning model that can analyze comments and classify them into three categories: positive("پیشنهاد میکنم"), negative("پیشنهاد نمیکنم"), and neutral("خنثی"). In general, the following sections of this article are as follows: Section 2 is a review of previous research in the field of sentiment analysis in Persian language. Section 3 on collected data. Section 4 covers data preprocessing, Section 5 covers models used, emotion classification tasks, score resampling methods, data pretreatment and segmentation, experimental design, evaluation criteria, and related results. . . A summary of the paper's conclusions and final thoughts is given in Section 6

## 2   Related Work

Early in the new millennium, Bo Pang and Lillian Lee[7]coined the term sentiment analysis. A natural language processing (NLP) approach called sentiment analysis, often known as opinion mining, aids in identifying the text's emotional undertone. Using sentiment analysis technologies, organizations may extract insightful information from unstructured and chaotic online content such as emails, blog posts, support tickets, web chats, social media channels, forums, and comments. Algorithms replace human data processing in rule-based, automated, or hybrid approaches. While rule-based systems utilize predefined lexicon-based rules to do sentiment analysis, automated systems use machine learning techniques to learn from data. Hybrid sentiment analysis makes use of both techniques [8].

There is presently a dearth of literature that focuses only on sentiment analysis in Persian literature, despite recent developments in deep learning-based approaches. Nonetheless, a few investigations have greatly expanded our knowledge in this field:

Deep learning techniques were used by Roshanfekr et al. [9] to construct a dataset of Persian electronic product reviews. Despite the excellence of their work, more investigation is required to completely comprehend this subject.In comparison to traditional Multi-Layer Perceptrons, Dashtipour et al.'s innovative approach produced an amazing 82.86% accuracy gain [10]. To evaluate a distinct dataset of Persian film critique, they combined Deep Auto Encoders with Deep CNNs.According to Nezad Bokaee et al.[11], they employed Word2Vec, CNNs, and LSTMs for word representation in a deep learning framework created especially for Persian sentiment analysis. Their model's accuracy rate was 85%.Studies on BERT pre-trained language models and attention mechanisms [12]outperformed CNN, LSTM, and skip-gram models in Persian sentiment analysis, despite the dearth of available data.Zobeidi and co-authors [13]developed a sentence-level opinion classifier with an amazing 95% accuracy rate using CNN and Word2Vec.Dastgheib et al. [14]introduced a hybrid methodology that combined CNN and SCL (Selective Cross-Language), which resulted in a notable gain of over 10% in sentiment categorization across diverse domains.Vazan, M., and Razmara, J. [15] jointly modelled aspect and polarity for aspect-based sentiment analysis in Persian reviews using a two-class classifier. An accuracy of 80% was attained by using group classification approaches such as CNN, LSTM, Bi-LSTM, and GRU.Sabri, N., Edalat, A., & Bahrak, B. [16]: Focusing on Persian-English code-mixed texts, their three-class classifier, which combined Bi-LSTM with attention mechanisms and BERT word embeddings, achieved an accuracy of 66.17% with an F-score of 63.66%.These studies collectively contribute to the growing body of work in deep learning-based sentiment analysis for Persian texts.

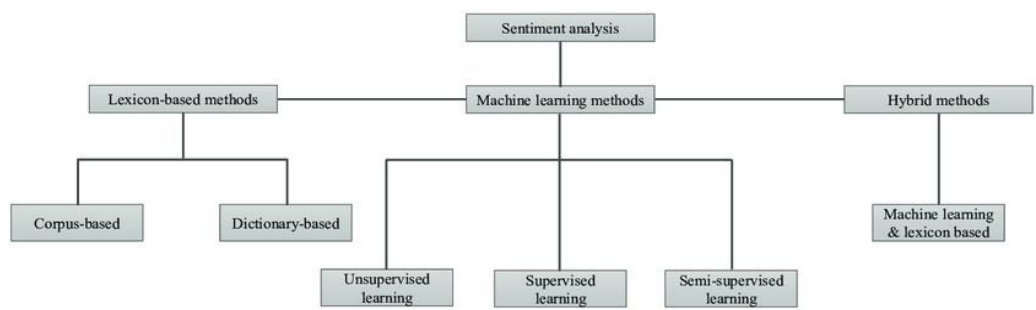

**Fig. 1.** Categories of sentiment analysis [8].

## 3 Data Set

### 3.1 Domains & Data collection

The PRFashion24 dataset, collected from comments on various large Persian online stores, was reviewed and gathered over a period of one month. As shown in Fig. [2], thoughts related to items such as clothes and wearables are included, with the data divided into three groups: women, men, and children (girls and boys), as seen in Fig. [3]. The dataset comprises 767,271 rows and 12 columns of data. To maintain the accuracy of the reviews, no changes or deletions were made to the data charts. However, further adjustments were made to improve the performance of the models on the dataset labels. The 204,570 people who participated in these polls were a unique sample, while the rest of the users shared their thoughts anonymously. Each review has a numerical score from 1 to 5. Eighteen different types of tags were finally divided into three categories in the data processing and cleaning stage. In the confusion matrix Fig. [5], the number of labels and their amounts can be seen in different categories. As a result, the evaluations and attitudes toward fashion sentiments are reflected in the name " PRFashion24."

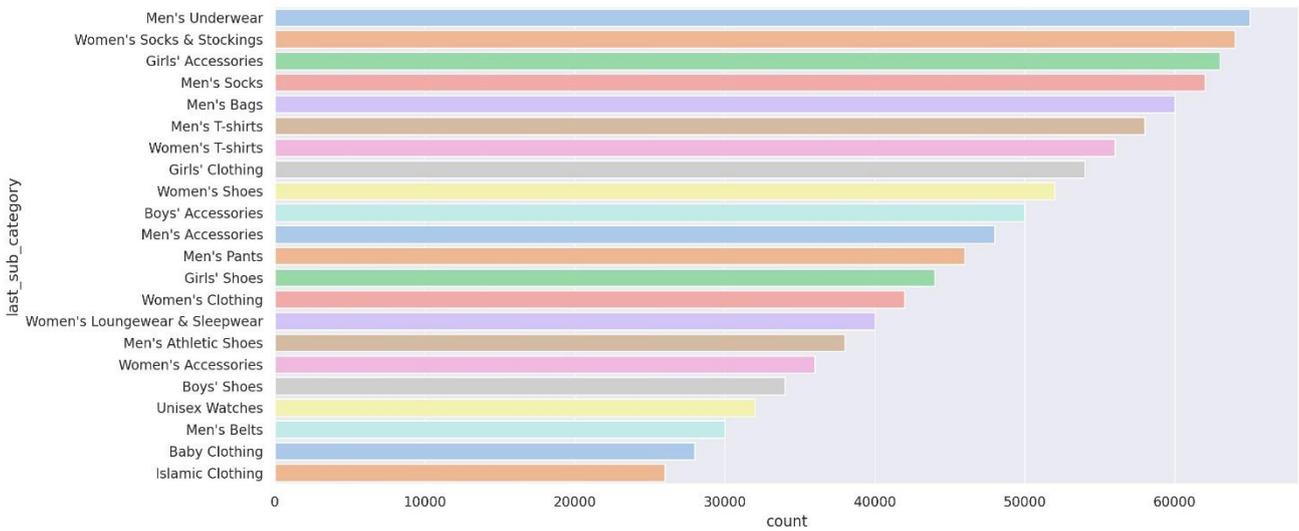

**Fig. 2.** Categories of sentiment analysis .

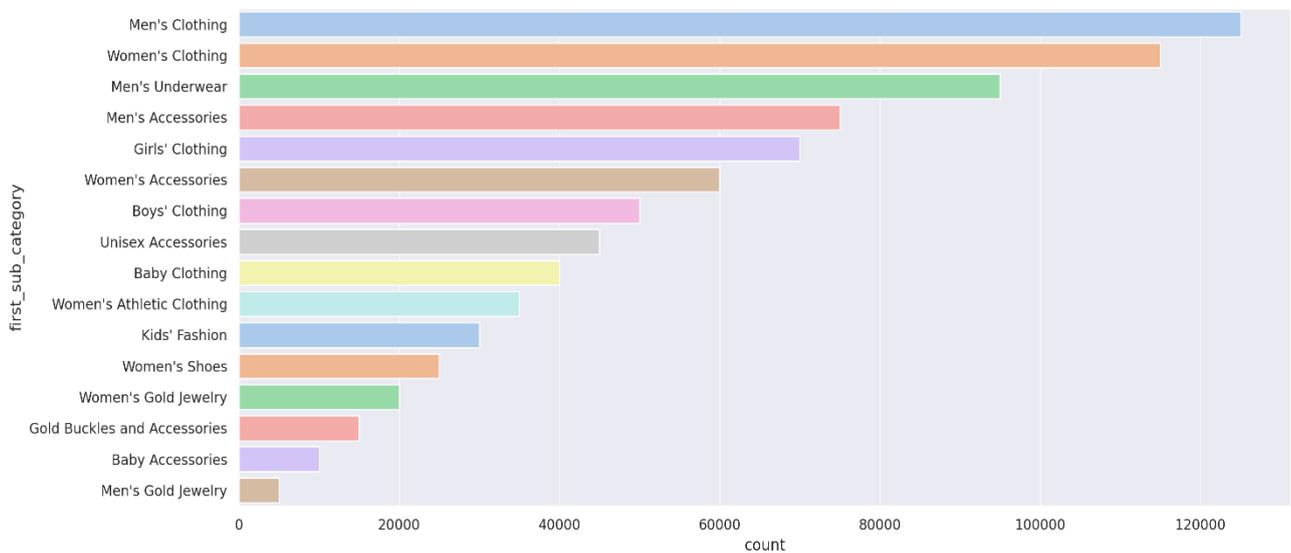

**Fig. 3.** Categories of sentiment analysis.

## 3.2 Data Visualization Tools

In this study, the patterns in the data were visually represented using a variety of graph formats, such as word clouds, line graphs, and bar graphs. For the purpose of visualization, many pre-built Python libraries were employed.

Persian hyperwords in Fig.[4] are created by grouping the most often occurring words from user comments together so that the most frequently occurring terms are bolded and larger.

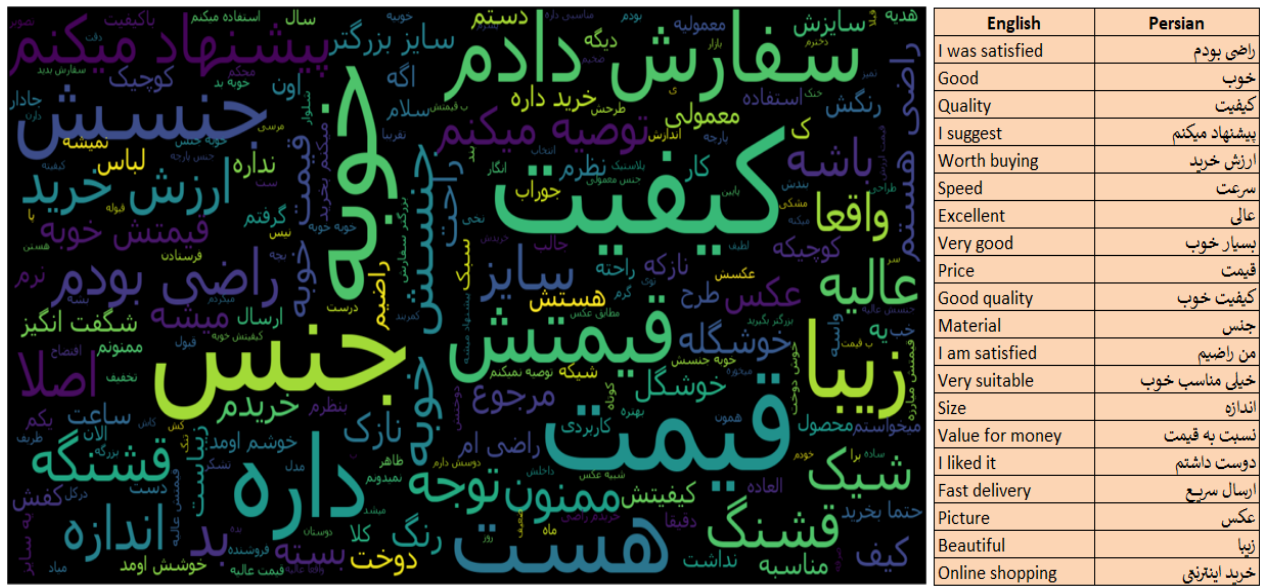

**Fig. 4.** Prevalent words in tweets .

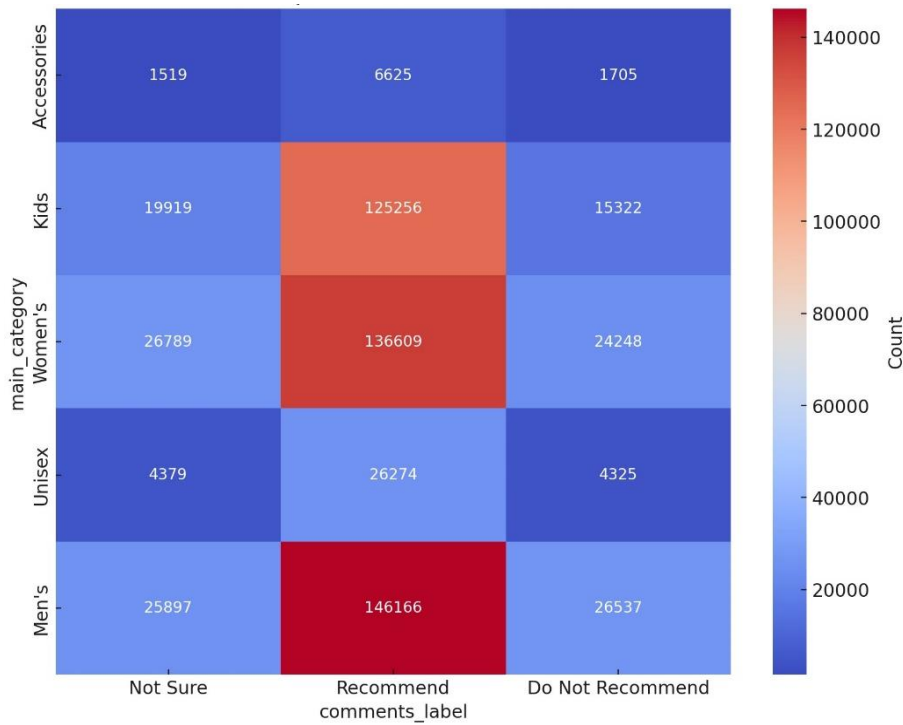

**Fig. 5.** Prevalent words in tweets.

## 4 Methodology

This section goes into great length on the suggested method for critiquing Persian viewpoints. The suggested framework is depicted in Fig.[6], and further information is given in the sections that follow.

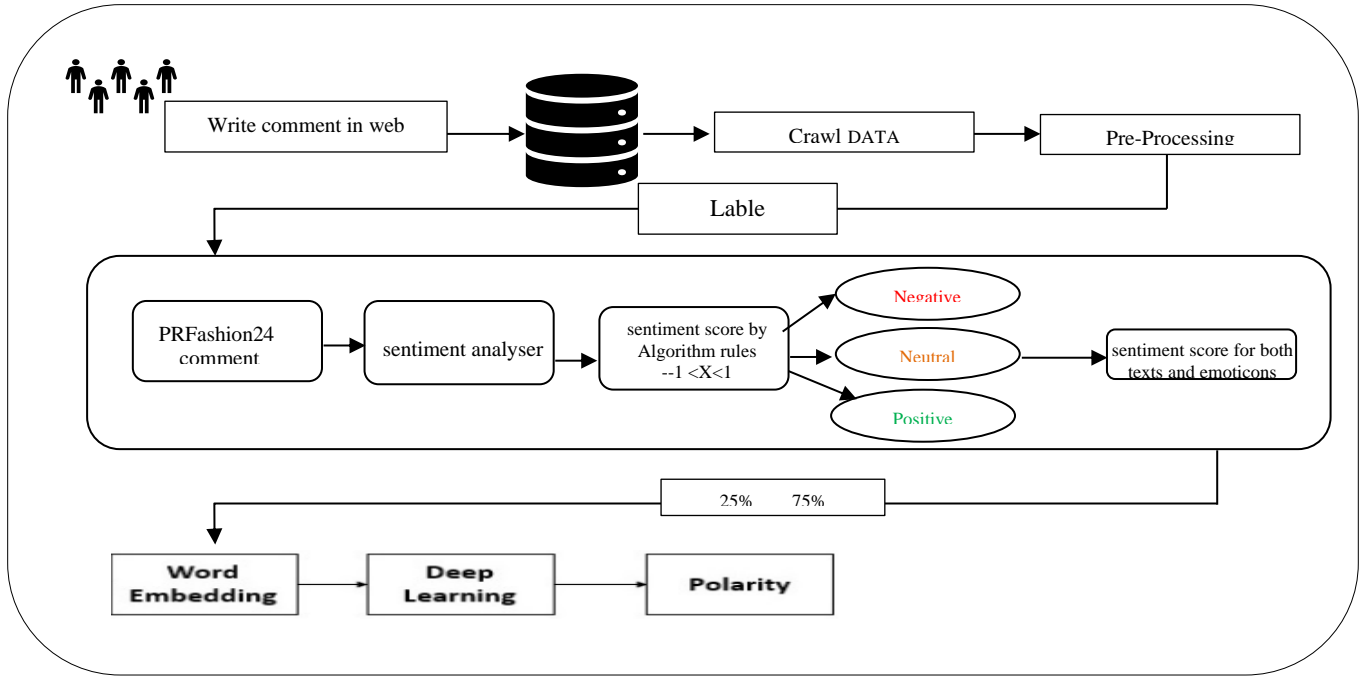

**Fig. 6.** Proposed Framework .

### 4.1 Data pre-processing

The PRFashion24 dataset, originally collected from various Persian online stores, initially contained 767,272 records. After removing comments with fewer than two words, more than 150 words, and middle lines without comments and tags, the dataset was reduced to 511,076 records. Subsequent steps involved extensive preprocessing: all comments were checked and cleaned by removing non-word characters, emojis, ZWNJ (Zero-width non-joiner) characters, non-Persian characters, sentence-ending punctuation, and extra spaces. After these steps, the comments were normalized.

Neural network training utilized a portion of this dataset, with 25% allocated for testing and validation and the remaining 75% for training. The dataset was labeled into three categories: neutral, negative, and positive. Preprocessing included tokenization, which involves splitting phrases into individual words based on normalization and punctuation rules. The NLTK tokenizer was used for tokenizing Persian expressions. For example, the phrase "لباس عالی است" ("لباس" (clothes), "عالی" (excellent), "است" (is)) is tokenized into "لباس" (clothes), "عالی" (excellent), and "است" (is). The normalization technique was employed to replace abbreviations with their actual meanings and to convert words to their literal forms. For instance, "من ازخریدم خوشحال هستم" ("I am happy with my purchase") is normalized to "خریدم خوشحالم" ("happy with my purchase"). Additionally, common prepositions such as 'و' ('and'), 'در' ('in'), 'به' ('to'), 'از' ('from'), 'که' ('that'), and 'این' ('this') were removed from the comments as stopwords. In Fig. [6], a portion of the dataset before preprocessing is depicted. The raw data includes Persian and English numbers, emojis, and English letters, all of which were removed during the data cleaning process.

**Fig. 6.** Dataset — Comments before processing.

# 5 Sentiment analysis model

## 5.1 Deep Neural Networks

To identify the orientation of users' opinions towards online shopping in the reviews in the PRFashion24 dataset, we used two different deep neural networks: an LSTM model and a hybrid BiLSTM-CNN model, as shown in Fig. [7]. We chose these networks due to their distinct text modeling capabilities. Specifically, a one-dimensional CNN can extract local features effectively, while a BiLSTM performs well in capturing long-range dependencies in textual information. The hybrid model leverages both CNN and BiLSTM architectures for complementary benefits.

In our architecture, a Conv1D layer with 64 one-dimensional convolutional filters is followed by a GlobalMaxPooling1D layer. The input layer receives a text comment with a fixed word length of 150, followed by an embedding layer for word embeddings with a dimension of 40. Subsequently, three Bidirectional LSTM (BiLSTM) layers are utilized, each with 128 units. Finally, a fully connected dense layer with a softmax activation function and three units is used to calculate the probability distribution across three sentiment orientations: positive, neutral, and negative.

The Table[1] outlines the architecture and parameters of our LSTM-based model. It begins with an Embedding layer that converts words into 20-dimensional vectors. This is followed by three LSTM layers with 128 units each, allowing the model to capture sequential dependencies in the text. The final LSTM layer outputs a 128-dimensional vector. A Dense layer with 128 units follows, acting as a fully connected layer to further process the features. The output layer is a Dense layer with 3 units and a softmax activation function, which computes the probability distribution across three sentiment classes: positive, neutral, and negative.

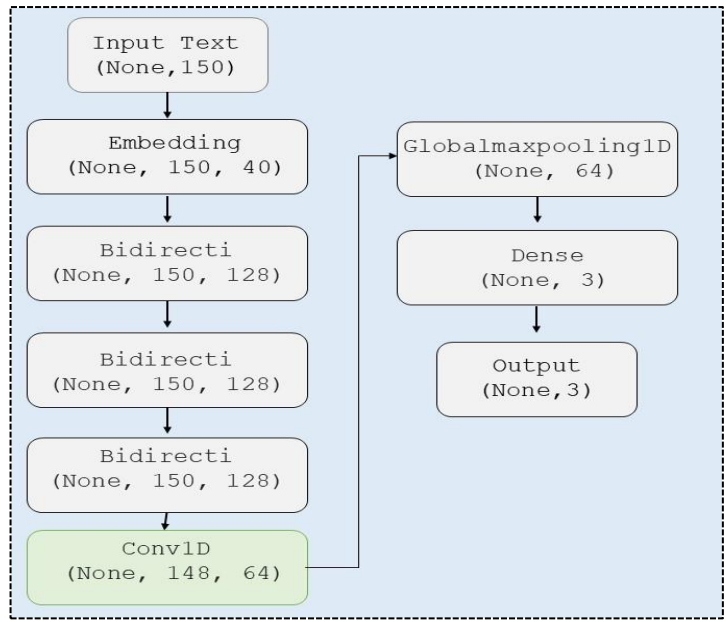

**Fig. 7.** Hybrid BiLSTM-CNN Architecture

```
Layer (type)                 Output Shape              Param #
=================================================================
embedding_1 (Embedding)      (None, None, 20)          20000

lstm_3 (LSTM)                (None, None, 128)         76288

lstm_4 (LSTM)                (None, None, 128)         131584

lstm_5 (LSTM)                (None, 128)               131584

dense_2 (Dense)              (None, 128)               16512

dense_3 (Dense)              (None, 3)                 387

=================================================================
Total params: 376355 (1.44 MB)
Trainable params: 376355 (1.44 MB)
Non-trainable params: 0 (0.00 Byte)
```

**Table. 1.** LSTM Architecture

## 5.2 Performance Metrics

The models' performance was assessed using a number of standard measures, including as accuracy (A), precision (P), recall (R), and F1-score (F1). Considering that the dataset is unbalanced and all classes bear equal weight, we chose macro-averaging, which is derived from the arithmetic (i.e., unweighted) mean of all F1-scores per class. This ensures that all classes receive the same treatment during the evaluation, and if the model performs worse on minority classes, it will result in a larger penalty[17] These formulas for binary classification are shown in Fig.[2] .

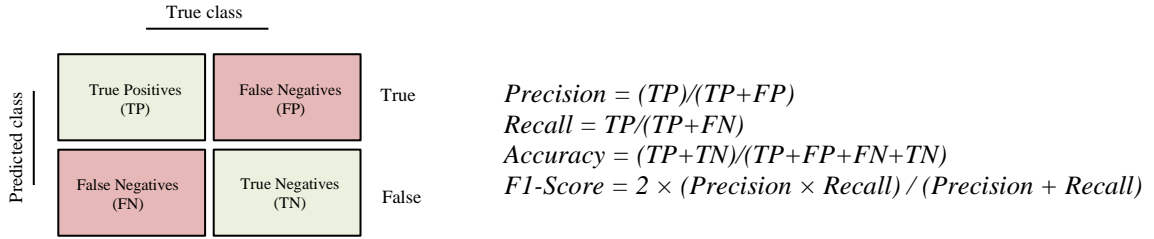

**Fig. 2.** Confusion matrix with the formulas of precision, recall, accuracy and f1-score

## 5.3 Performance Model

In the next step, we apply the performance metrics outlined in Table[2] to assess the effectiveness of our models, a crucial phase in our analysis using the PRFashion24 dataset. We examine the capabilities of two deep learning architectures: LSTM (Long Short-Term Memory) and BiLSTM-CNN (Bidirectional Long Short-Term Memory with Convolutional Neural Network). Our comparative analysis underscores the enhanced performance of the BiLSTM architecture, attributed to the integration of a bidirectional network with a convolution layer.

These findings are briefly summarized in Table [1] .The F1 score emerges as a fundamental criterion when examining emotion classification accurately. In particular, the LSTM model achieved an F1 score of 81%, while the BiLSTM-CNN model surpassed this score with an F1 score of 82%, representing a significant improvement of 1%. This development emphasizes the effectiveness of the BiLSTM-CNN model in capturing complex patterns in datasets, thereby facilitating more accurate sentiment classification.

| class | LSTM | | | Bilstm-CNN | | |
|---|---|---|---|---|---|---|
| | P | R | F1 | P | R | F1 |
| Neutral | 0.44 | 0.26 | 0.33 | 0.47 | 0.27 | 0.34 |
| Negative | 0.67 | 0.62 | 0.64 | 0.69 | 0.63 | 0.66 |
| Positive | 0.87 | 0.94 | 0.94 | 0.87 | 0.95 | 0.91 |
| Weighted avg | 0.87 | 0.81 | 0.79 | 0.79 | 0.82 | 0.80 |

**Table.2.** Performance of 1D-CNN w/o attention.

## 6 Conclusions and Future Work

This study was conducted to explore the sentiments of Farsi-speaking consumers towards the online shopping experience on various websites, utilizing the PRFashion24 dataset, which includes 767,272 reviews collected from multiple online stores. The initial data preprocessing stages—cleaning, tokenization, stop word removal, and data normalization—were meticulously carried out in two phases. Our research showcases the use of deep learning methods for sentiment analysis, employing basic Natural Language Processing (NLP) techniques to categorize emotional sentiments into three distinct polarities: positive, negative, and neutral.

Our analysis, as of May 2024, revealed that 70% of respondents expressed positive sentiments, 11% negative, and 19% neutral. The study employed LSTM and BiLSTM-CNN architectures to assess their prediction accuracy. These models demonstrated proficiency in sentiment prediction and analysis, with the LSTM architecture achieving an accuracy of 81.23%, and the BiLSTM-CNN model showing an enhanced accuracy of 82.85%. Both models exhibited commendable performance in terms of precision, recall, F-1 scores, and confusion matrix analysis. Looking forward, our future research will aim to expand the applicability of our models by testing with alternative datasets such as those derived from Twitter and by exploring additional sentiment analysis tools to further enhance the robustness and accuracy of our findings.